\documentclass[letterpaper, 10 pt, conference]{ieeeconf}  % Comment this line out if you need a4paper
\IEEEoverridecommandlockouts                              % This command is only needed if 
\usepackage{graphics} % for pdf, bitmapped graphics files
\usepackage{graphicx} % for includegraphics
\usepackage{times} % assumes new font selection scheme installed
\usepackage{amsmath} % assumes amsmath package installed
\usepackage{amssymb}  % assumes amsmath package installed
\usepackage{url}
\usepackage{hyperref}
\usepackage{xcolor}
\usepackage{bm}
\usepackage{xspace}
\usepackage{subcaption}

\definecolor{overviewpurple}{RGB}{148,103,189}
\definecolor{overviewgreen}{RGB}{44,160,44}
\definecolor{overviewblue}{RGB}{31,119,180}
\definecolor{overviewred}{RGB}{214,39,40}

\usepackage[draft]{changes}
\definechangesauthor[name={Mederic}, color=teal]{MF}
\definechangesauthor[name={Jan}, color=orange]{JB}
\definechangesauthor[name={Vladimir}, color=brown]{VP}
\definechangesauthor[name={JosefReview}, color=blue]{VPR}
\definechangesauthor[name={Josef}, color=teal]{JS}

\DeclareMathOperator*{\argmin}{arg\,min}
\newcommand{\norm}[1]{\left\lVert#1\right\rVert}
\DeclareMathOperator{\diag}{diag}

\newcommand{\ftracker}{\relax\ifmmode f_\text{track}\else $f_\text{track}$\fi}
\newcommand{\flocalize}{\relax\ifmmode f_\text{localize}\else $f_\text{localize}$\fi}
\newcommand{\folt}{\relax\ifmmode f_\text{OLT}\else $f_\text{OLT}$\fi}

\makeatletter
\DeclareRobustCommand\onedot{\futurelet\@let@token\@onedot}
\def\@onedot{\ifx\@let@token.\else.\null\fi\xspace}

\def\eg{\emph{e.g}\onedot} 
\def\ie{\emph{i.e}\onedot}

\makeatother

\title{\LARGE \bf Visually Guided Model Predictive Robot Control via \\ 6D Object Pose Localization and Tracking }

\author{Mederic Fourmy$^{\clubsuit}$ \and Vojtech Priban$^{\clubsuit}$ \and Jan Kristof Behrens$^{\clubsuit}$ \and Nicolas Mansard$^{\diamondsuit}$ \and Josef Sivic$^{\clubsuit}$ \and Vladimir Petrik$^{\clubsuit}$%
\thanks{$^{\clubsuit}$ CIIRC, Czech Technical University in Prague}%
\thanks{$^{\diamondsuit}$ LAAS-CNRS, Universite de Toulouse, CNRS, Toulouse}%
\thanks{This work was partly supported by the AGIMUS project, funded by the European Union under GA no.101070165, by the European Regional Development Fund under project Robotics for Industry 4.0 (reg. no. $CZ.02.1.01/0.0/0.0/15\_003/0000470$), and by the Czech Science Foundation (project no. GA21-31000S). Views and opinions expressed are however those of the author(s) only and do not necessarily reflect those of the European Union or the European Commission. Neither the European Union nor the European Commission can be held responsible for them.}%
}

\begin{document}

\maketitle
\thispagestyle{empty}
\pagestyle{empty}

\begin{abstract}
The objective of this work is to enable manipulation tasks with respect to  the 6D pose of a dynamically moving object using a camera mounted on a robot. Examples include maintaining a constant relative 6D pose of the robot arm with respect to the object, grasping the dynamically moving object, or co-manipulating the object together with a human. Fast and accurate 6D pose estimation is crucial to achieve smooth and stable robot control in such situations. 
The contributions of this work are three fold. 
First, we propose a new visual perception module that asynchronously combines accurate learning-based 6D object pose localizer and a high-rate model-based 6D pose tracker. The outcome is a low-latency accurate and temporally consistent 6D object pose estimation from the input video stream at up to 120~Hz.  
Second, we develop a visually guided robot arm controller that combines the new visual perception module with a torque-based model predictive control algorithm. Asynchronous combination of the visual and robot proprioception signals at their corresponding frequencies results in stable and robust 6D object pose guided robot arm control.    
Third, we experimentally validate the proposed approach on a challenging 6D pose estimation benchmark and
 demonstrate 6D object pose-guided control with dynamically moving objects on a real 7~DoF Franka Emika Panda robot.

\end{abstract}

\section{Introduction}

\begin{figure}[t]
    \centering
    \includegraphics[width=\linewidth]{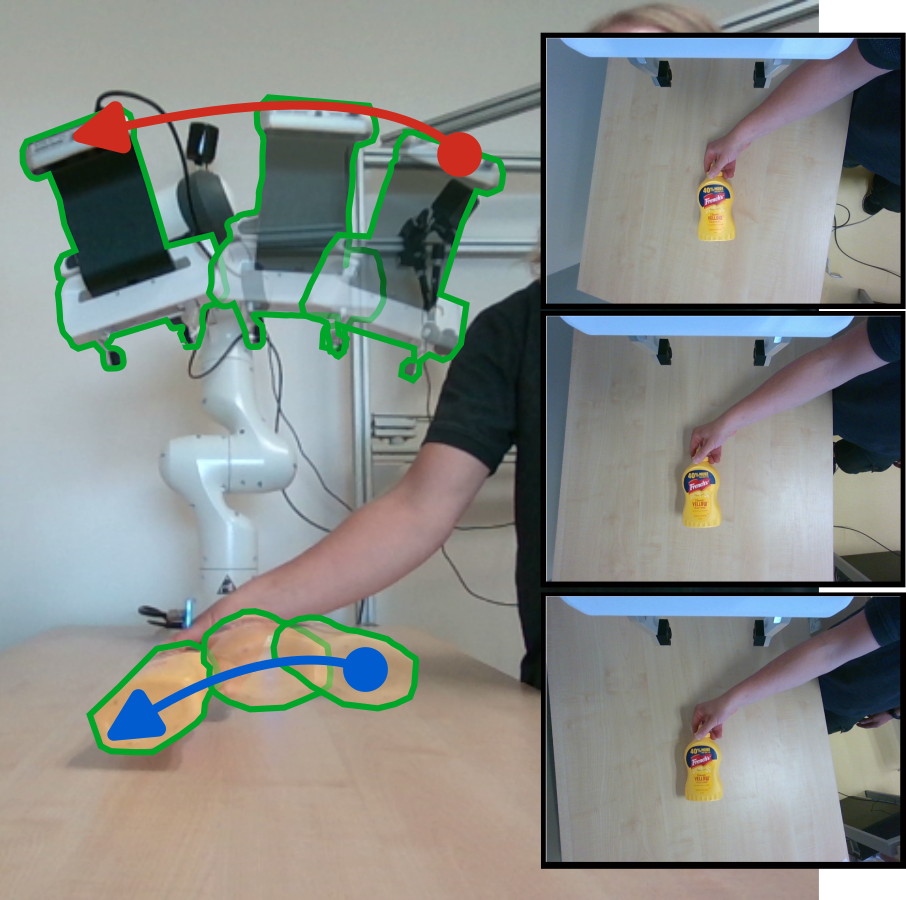}
    \caption{
    \textbf{Robot arm control by 6D pose of the object.
    % Object 6D pose tracking.
    }
    The objective is to control the robot arm with a mounted camera (\textcolor{overviewred}{red arrow}) by commanding joint torques such that the object 6D pose (\textcolor{overviewblue}{blue arrow}) w.r.t. the camera remains constant.
    This is illustrated by three frames (see insets) captured by the robot camera corresponding to the robot/object poses  \textcolor{overviewgreen}{shown by green contours in the main image}. 
    Please note (see the insets) how the object pose remains stable while the background changes in the captured frames.
    \textbf{More results and experimental analysis in the companion video.}
    \vspace{-0.5cm}
    }
    \label{fig:teaser}
\end{figure}

Visually-guided control is at the core of many robotic applications, from path following by mobile robots \cite{furgale2010visual} to visual servoing \cite{chaumette2016visual}. In order to achieve a stable and robust feedback loop, the perception system has to recover the estimated state both accurately and at a high rate. In the context of object manipulation, a commonly chosen state representation is the 6D pose of objects of interest, \ie, the 3D translation and 3D rotation of the objects in the scene with respect to the camera coordinate frame. 
While some manipulation tasks can be achieved with a static scene model \cite{chabal2022assembly}, 
many applications are of an inherently dynamic nature with human-robot handovers~\cite{Ortenzi_Cosgun_Pardi_Chan_Croft_Kulic_2022, Salehian18}, human-robot co-manipulation~\cite{Figueroa20} or mobile manipulation~\cite{Haviland_Sünderhauf_Corke_2022} being the prime examples. 
This is challenging as it requires accurate and low-latency 6D pose estimation of the target objects in the scene.
In addition, pose estimates need to be integrated with a robust and reactive controller that is capable of meeting the dynamic requirements of the application.

Despite promising recent progress in object detection and 6D pose estimation \cite{hodavn2020bop, sundermeyer2023bop, labbe2020cosypose, haugaard2022surfemb, lipson2022coupled, labbe2022megapose}, 6D object pose estimation algorithms usually focus on accuracy rather than speed. On the other end, object 6D pose tracking methods offer fast pose updates of already detected objects, but require an initialization from the user~\cite{stoiber2022iterative}. As a result of these limitations, many real-world applications rely instead on fiducial markers~\cite{fiala2005artag,garrido2014automatic,olson2011apriltag}, motion capture systems~\cite{Figueroa20,Salehian18} or ad-hoc detection such as color based segmentation~\cite{parosi2023kinematically}. 

In this paper, we propose a visual perception module that builds on (\textit{i})~state-of-the-art accurate learning-based 6D object pose detector and (\textit{ii})~state-of-the-art high-rate model-based 6D pose tracker to achieve object pose estimation limited only by the rate of image acquisition. 
Further, we develop a visually guided robot controller based on the model predictive control (MPC) that is able to reactively incorporate perception updates to meet application targets (\eg, positioning the robot's gripper). The \textit{anticipatory} nature of MPC makes it particularly suitable for generating efficient motions in real time~\cite{dantec2022whole}.
In principle, the only input data required is a 3D object model to configure the 6D pose tracker and the 6D pose estimator. Our method runs the pose detector and several instances of the object tracker in separate processes and uses the results of the pose detector to re-initialize the trackers. New images are directly fed to a tracker so that a pose estimation result is available within the tracker runtime of approximately $5$~ms.
To quantitatively validate our perception module, we build on the BOP challenge evaluation~\cite{sundermeyer2023bop} and the YCBV dataset~\cite{calli2015benchmarking} to measure the performance of the proposed method and several baseline methods.
Lastly, we demonstrate visually-guided robot arm control with hand held objects (see Fig.~\ref{fig:teaser}). 

 \noindent\textbf{Contributions.}
 The paper has the following three main contributions:
 (\textit{i})~we propose a new visual perception module that asynchronously combines accurate learning-based 6D object pose localizer and a high-rate model-based 6D pose tracker;
 (\textit{ii})~we develop a visually guided robot arm controller that leverages the new visual perception module in a torque-based model predictive control algorithm; and
 (\textit{iii})~we experimentally validate the proposed approach on a challenging 6D pose estimation benchmark and
 demonstrate 6D object pose-guided control with dynamically moving objects on a real 7~DoF Franka Emika Panda robotic platform.
 We will make the code publicly available.

%%%%%%%%%%%%%%%%%%%%%
\section{Related work}\label{sec:rw}
\noindent\textbf{Object 6D localization.}
The field of object detection and 6D pose estimation has shown impressive progress over recent years, which has been documented and fostered by the \textit{benchmark for 6D object pose estimation }(BOP) and the associated BOP challenge~\cite{hodavn2020bop,sundermeyer2023bop}. 
Most methods follow a two-step approach, first performing object detection in RGB frames~\cite{he2017mask} followed by 6D pose estimation, assuming the availability of object meshes.
Learning-based techniques have dominated the field.
Among the diverse approaches, render-and-compare methods~\cite{li2018deepim,labbe2020cosypose,labbe2022megapose}, have achieved superior performance by iteratively refining the 6D pose based on predictions by a neural network.
Due to their iterative nature these methods achieve superior performance but are slow to be used in real-time control.
In this work, we propose to combine a slow "render and compare" 6D pose localizer with a fast 6D pose tracker.
Although the proposed approach can work with an arbitrary localizer, we use pre-trained CosyPose~\cite{labbe2020cosypose} in all our experiments. 

\noindent\textbf{Object 6D pose tracking.}
When a good initial guess of the object 6D pose is available, it can be tracked frame to frame by fast local methods.
Object pose tracking methods rely on object edges~\cite{harris1990rapid}, extracted point features~\cite{rosten2005fusing,comport2006real}, or depth~\cite{trinhmodular}. 
Region-based tracking approaches
% , which draws its inspiration from level-set segmentation~\cite{cremers2007review,bibby2008robust}, 
propose to solve the 2D object shape segmentation and 6D pose tracking problems jointly by constructing an image-wise posterior distribution~\cite{rosenhahn2007three}. Although initial versions required highly optimized GPU implementations to run at the camera frequency~\cite{prisacariu2012pwp3d,prisacariu2014real}, a sparse formulation based on contour point sampling~\cite{kehl2017real} dramatically reduces the computation time, down to a few milliseconds per image \cite{stoiber2022srt3d,stoiber2022iterative} on a single CPU.
In this work, we rely on the ICG implementation~\cite{stoiber2022iterative} that features both region-based and depth modalities.
The combination of the 6D pose localizer and the local 6D pose tracker that we propose in this paper combines the benefits of the both world, \ie~the detection capability and the accuracy of the localizer and speed of the tracker.

\noindent\textbf{Visual servoing.}
Visual servoing aims at building a closed-loop controller using visual information from a camera stream to achieve a certain goal. The various methods are traditionally classified into two broad categories~\cite{chaumette2016visual}:
(\textit{i}) image-based visual servoing~\cite{weiss1987dynamic,feddema1989vision}, which uses 2D geometric primitives (\eg, points, curves) to define control objective in an image space;
and (\textit{ii}) pose-based visual servoing~\cite{wilson1996relative,thuilot2002position} which assumes the availability of the estimate of a target 6D pose.
Some works propose switching between image- and pose-based servoing depending on the phase of movement~\cite{haviland2020control}.
Although image-based servoing allows to naturally incorporate visibility constraints in the control law, pose-based servoing is closer to applications such as object grasping~\cite{Salehian18}. We address the challenge of obtaining robust and fast estimates of 6D poses for pose-based control.
These works typically implement control laws at the joint velocity level, which lack the natural impedance of torque-based control.

\noindent\textbf{Model predictive control for visual servoing.}
For image-based visual servoing, MPC has shown superior performance to traditional reactive controllers~\cite{sauvee2006image,allibert2009can}.
System dynamics in an image space is either approximated analytically~\cite{sauvee2006image,allibert2009can} or computed through learning-based optical flow estimation methods~\cite{katara2021deepmpcvs}.
Image-based MPC servoing was successfully applied to control drone~\cite{jacquet2020motor}, legged platform~\cite{parosi2023kinematically}, or mobile manipulator~\cite{bildstein2022visual, bildstein2023multi}.
Contrary to the state-of-the-art methods using image space MPC, we propose using MPC for pose-based control, where 6D pose is obtained by the proposed perception module.

\begin{figure*}[t]
    \centering
    \includegraphics[width=\linewidth]{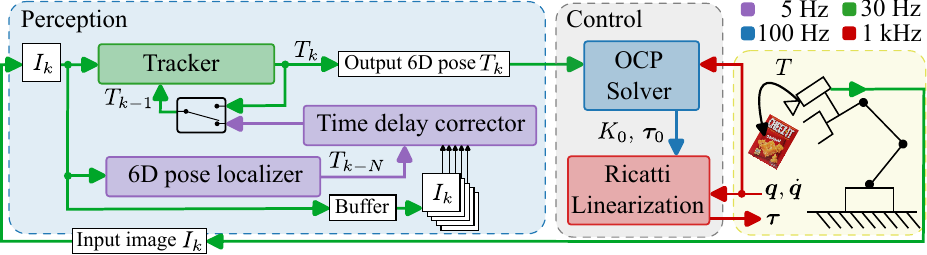}
    \caption{
    \textbf{Overview of the perception-control cycle.}
    The objective of the feedback control is to track 6D pose of an object seen by a camera, as illustrated on the right by a robot and red cheez-it box.
    To achieve that, we designed a perception module that runs a fast local \textcolor{overviewgreen}{Tracker}  on an input image~$I_k$ with the initial pose~$T_{k-1}$ selected either from the previous run of the tracker or from the \textcolor{overviewpurple}{6D pose localizer \& Time delay corrector} modules, if that information is available.
    The \textcolor{overviewpurple}{6D pose localizer} is slow and
    the objective of the \textcolor{overviewpurple}{Time delay corrector} is to \textit{catch-up} in time by quickly tracking through images stored in the buffer while the 6D pose localizer was computing.
    The output of the tracker, the pose~$T_k$, is used by the \textcolor{overviewblue}{OCP solver} to compute Ricatti gains~$K_0$ and torques~$\bm \tau_0$ that are used by the \textcolor{overviewred}{Ricatti Linearization} module to provide fast feedback for real-time robot control.
    Typical processing frequencies of individual modules are \textcolor{overviewpurple}{5~Hz} for the 6D pose localizer and the time delay corrector, \textcolor{overviewgreen}{30~Hz} for the camera and tracker, \textcolor{overviewblue}{100~Hz} for the OCP solver, and \textcolor{overviewred}{1~kHz} for real-time robot control.
    }
    \label{fig:overview}
\end{figure*}

\section{Object pose guided model predictive control}\label{sec:method}
The objective of the proposed system is to perform manipulation tasks with respect to  the 6D pose of a dynamically moving object using a camera mounted on a robot.
The key technical challenge in such situations addressed by our method lies in achieving an accurate 6D pose estimation without introducing a significant delay into the control loop.
The estimated 6D pose is then used in combination with MPC to achieve optimal control of the robot.

The proposed method is a 1~kHz torque level MPC controller taking reference from 6D object poses obtained from the 30~Hz image stream.
To achieve this real-time robot control performance, the perception and control modules run asynchronously, as shown in Fig.~\ref{fig:overview}.
The perception module detects objects of interest in the scene and tracks them in a fast and temporally consistent manner as described in Sec.~\ref{sec:perception}.
The key innovation is handling the inherent asynchronicity of the accurate-but-slow 6D pose localizer and fast-but-local tracker via the \textit{time delay corrector} module that operates on the buffered images in order to \textit{catch-up} in time.  
The 6D poses of objects detected by the perception module are used by the Optimal Control Problem (OCP)~\cite{mastalli2020crocoddyl} solver and a feedback controller using a Ricatti Gains linearization of the solution~\cite{dantec2022first} to compute torques for the robot at 1~kHz, as described in Sec.~\ref{sec:control}.
In the following sections, $I$ symbols represent RGB image frame, while $T \in \mathbf{SE}(3)$ is a rigid body transformation.

\subsection{Temporally consistent 6D object pose tracker}\label{sec:perception}
The objective of the perception module is to compute the 6D poses of objects in the scene based on the input image~$I_k$, observed at discrete time $k$ while introducing as little delay as possible.
Although the proposed method can track an arbitrary number of objects, we describe the tracking of a single object pose~$T_k$ to simplify the notation.

\noindent\textbf{6D pose localizer.}
With unlimited computational resources, the pose of an object can be estimated by the 6D pose localizer~$T_k = \flocalize (I_k)$ that detects the object of interest in the image and estimates its 6D pose with respect to the camera coordinate frame and, as a consequence, also the robot coordinate frame, as we assume the camera is calibrated with respect to the robot. 
However, robust object localizers are slow, \eg~$0.25$~s for CosyPose~\cite{labbe2020cosypose} or even $30$~s for MegaPose~\cite{labbe2022megapose} on up-to-date hardware (see Sec.~\ref{sec:experiments}).
The long computation time makes the localizer impractical for closed-loop control.

\noindent\textbf{Tracker.}
To mitigate this limitation, we combine the localizer with a fast local tracker~$T_k = \ftracker (I_k, T_{k-1})$ that computes the pose of objects from a given image~$I_k$ and initial guess of the pose~$T_{k-1}$.
Compared to the localizer, a single pass of the tracker is fast, introducing only a few milliseconds delay into the system.
However, it acts only locally, and it thus requires an initial pose that is refined based on the observed image.
The tracker is not able to discover the presence of new objects, nor does it detect that the object is no longer visible by the camera when it is, for example, occluded.

\noindent\textbf{Object localization and tracking (OLT).}
We combine the localizer and the tracker into a single perception module~$T_k = \folt (I_k, T_{k-1})$ that computes fast feedback at the frequency of~$\ftracker$ while running~$\flocalize$ in parallel for object (re-)discovery and more accurate pose estimation.
Our architecture, shown in the perception plate of Fig.~\ref{fig:overview}, runs~$\ftracker(I_k, T_\text{init})$ on the current image with the initial pose~$T_\text{init}$ selected either from (i)~the previous iteration of the tracker, \ie~$T_{k-1}$ or (ii)~the separate process that localizes the object if that information has already been computed by the \textit{time delay corrector} for the previous image~$I_{k-1}$, \ie~end of the image buffer.
% We assume that the initial pose $T_0$ is computed in advance of tracking by running the localizer on the initial frame $I_0$, \ie~$T_0 = \flocalize (I_0)$.
The main tracker is initialized once the first frame to enter the system has been processed by the localizer and time delay corrector.

\begin{figure}
    \centering
    \includegraphics[width=\linewidth]{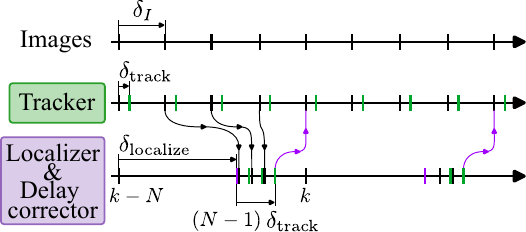}
    \caption{
    \textbf{Perception module timeline.}
    The first row illustrates the stream of images with typical delay between images $\delta_I$ being 33~ms.
    The second row illustrates the delay caused by the \textcolor{overviewgreen}{tracker} module (\ie~\ftracker), denoted by $\delta_\text{track}$ that corresponds to a few milliseconds and therefore output poses (\textcolor{overviewgreen}{green ticks}) are produced at the frequency of the input image stream.
    The tracker needs initial pose that is taken either from previous run of the tracker or from the  \textcolor{overviewpurple}{6D pose localizer \& time delay corrector} modules if possible as indicated by \textcolor{overviewpurple}{purple arrows}.
    The 6D pose localizer runs \flocalize{} (with typical $\delta_\text{localize}$ being a few hundreds of milliseconds) followed by \ftracker{} applied $N-1$ times on the buffered images (\textcolor{overviewgreen}{green ticks in the third row}).
    }
    \label{fig:timeline}
\end{figure}

\noindent\textbf{Time delay corrector.} 
In the parallel process, a single instance of the localizer is run all the time the resources are available.
Let us assume that the localizer started processing input image~$I_{k-N}$ at time $k-N$.
It takes some time to get output of~$\flocalize$ during which new images arrive and are stacked inside a buffer.
Once the pose $T_{k-N} = \flocalize(I_{k-N})$ is computed, a second instance of the tracker is run on all images inside the buffer, \ie
$T_{i} = \ftracker (I_i, T_{i-1})$ iteratively for $i \in \{ k-N+1, \, \ldots, \,  k-1 \}$ while providing the final pose computed at the time $k-1$ to the main tracker process.
The timeline of the perception module is illustrated in Fig.~\ref{fig:timeline}.
Note that our architecture assumes that the frequency of $\ftracker(\cdot)$ is higher than the frequency of the input image stream.
Otherwise, the localizer process would never \textit{catch-up} with the main process.
With the state-of-the-art tracker~\cite{stoiber2022iterative}, this feedback can be easily calculated for image frequencies up to 120~Hz.
However, the higher the input image frequency, the longer it takes to inject information from the localizer process.
This affects the tracking accuracy as we analyze in Sec.~\ref{sec:experiments}.

\subsection{6D pose-based visual servoing using MPC}\label{sec:control}
We design a controller that brings the camera attached to the robot end-effector to a user-defined relative pose w.r.t. the object pose~$T_k$ obtained from the object tracker.
As a practical application, one may choose a reference pose from a set of predefined grasp poses for a given object.
The challenge of the control lies in a real-time requirement of the robot to receive torque commands at 1~kHz.
To address this challenge, we build on~\cite{dantec2022first} and split the control into solving the optimal control problem at 100~Hz and computing 1~kHz feedback through Ricatti linearization, but here incorporating the 6D object poses as a guiding reference in the problem formulation.

\noindent\textbf{OCP solver.}
The control module's main objective is to follow the object given the latest object pose estimates $T_k$ provided by the perception and the current robot state $\bm x = \begin{pmatrix} \bm q^\top & \bm{\dot q}^\top \end{pmatrix}^\top$, with $\bm q$ and $\dot{\bm q}$ being the measured joint angles and velocities, respectively. We control the manipulator at the torque level (\ie~$\bm u = \bm \tau$) to be able to exploit the natural dynamics of the manipulator and obtain smoother motions.
Solving the Optimal Control Problem (OCP) produces optimal state and control trajectories over a fixed time horizon, where optimality is defined by a set of weighted high-level objectives. The resolution of the OCP is done in the framework of Differential Dynamic Programming (DDP)~\cite{mayne1966second} and is implemented using the Feasibility-driven DDP solver~\cite{mastalli2020crocoddyl}. The OCP is transcribed to a nonlinear program by discretizing the continuous problem using a direct multiple-shooting strategy:
\begin{align}
\begin{split}
    \argmin\limits_{\substack{~~\bm u_0, \ldots, \bm u_{M-1} \\ \bm x_1, \ldots, \bm x_M}} &\sum_{i=0}^{M-1} l_i(\bm x_i, \bm u_i) + l_M(\bm x_M)  \, , \\
    \text{s.t.} \quad \bm x_{i+1} &= f(\bm x_i, \bm u_i), \, \forall i \in \{0,\ldots, M-1\}, \\
    \bm x_0 &= \bm \hat{\bm x} \, ,
\end{split}\label{eq:OCP}
\end{align}
where $\bm{\hat{x}}$ is the latest robot state measurement, 
$\bm x_i$ and $\bm u_i$ are the state of the robot and the applied control at discrete time~$i$,
$f(\bm x_i, \bm u_i)$ describes the robot dynamics (\ie~articulated body algorithm),
and $l_i$ and $l_M$ are running and terminal costs, respectively.
The tracking objective of the control is specified by the costs, as we describe in Sec.~\ref{sec:costs}.
Given an initial guess, the DDP algorithm solves ~\eqref{eq:OCP} and returns a sequence of states and control actions by iterating Bellman recursions (see~\cite{mastalli2020crocoddyl} for more details).

\noindent\textbf{Ricatti linearization.}
Besides trivial systems and a short time horizon, it is impossible to solve OCP at the robot control frequency, \ie~1~kHz.
However, it has been shown~\cite{dantec2022first} that the Ricatti gains~$K_0$ obtained as a byproduct of the OCP solution can be used to implement a first-order approximation of the optimal policy. 
Denoting by $\bm \tau_0 = \bm u_0^\ast$ the first step of the optimal control obtained from the OCP solver, the linear approximation of the optimal policy is:
\begin{equation}
    \bm \tau (\bm x) = \bm \tau_0 + K_0(\bm x - \bm x_0) \, ,
    \label{eq:ricatti}
\end{equation}
where $\bm x$ is the latest robot state measurement and
$\bm x_0$ is the state of the robot for which the OCP solution was found.
This computation is immediate, it decouples the OCP problem complexity from the real-time constraints, and it allows us to solve long-horizon problems.

\subsection{Tracking objective}\label{sec:costs}
% \noindent\textbf{OCP cost objectives.}
The behavior of the controller is defined by the formulation of the running and terminal costs in eq~\eqref{eq:OCP}.
For our tracking problem, we formulate the costs as follows:
\begin{align}
\begin{split}
    l_i(\bm x_i, \bm u_i) &= w_v l_v(\bm x_i) + l_x(\bm x_i) + l_u(\bm x_i, \bm u_i) \, , \\
    l_M(\bm x_M) &= w_v l_v(\bm x_M) + l_x(\bm x_M)  \, ,
\end{split}\label{eq:costs}
\end{align}
where
$l_v(\cdot)$ is the tracking cost scaled by $w_v$ and $l_x(\cdot)$ and $l_u(\cdot)$ are state and control regularization costs, respectively.

\noindent\textbf{Tracking cost.}
We define a tracking cost to minimize the $\mathbf{SE}(3)$ distance between the estimated pose of the object and the reference pose of the object~$T_\text{ref}$, both expressed in the robot base frame, \ie: 
\begin{align}
 l_v(\bm x) = \norm{\log\left( \left( T_\text{BC}(\bm q_k) T_k \right)^{-1}  T_\text{BC}(\bm q) T_\text{ref} \right) }^2 \, ,
\end{align}
where 
$T_k = \folt(I_k)$ is object pose estimated by the proposed tracker, 
$T_\text{BC}(\cdot)$ represents the forward kinematics from the robot base to the camera,
and the operator $\log$ represents the $\mathbf{SE}(3)$ log map~\cite{sola2018micro}.
The tracking cost approaches zero if the transformation between the robot camera and estimated object pose approaches~$T_\text{ref}$.

\noindent\textbf{State regularization cost.}
We define the cost of state regularization as
$l_x(\bm x) = \left(\bm x- \bm x_\text{rest}\right)^\top Q_x \left(\bm x- \bm x_\text{rest}\right)$
with $\bm x_\text{rest}= \begin{pmatrix} \bm q_\text{rest}^\top & \bm{0}^\top \end{pmatrix}^\top$ penalizing joint configurations far from a fixed rest configuration~$\bm q_\text{rest}$ and penalizing high joint velocities at the same.
The objective of the regularization cost is to prevent robot \textit{null-space} motion, \ie~motion that does not affect the pose of the camera itself.
It is required for redundant robots, where the number of DoF for robot is higher than the task-space number of DoF.
Regularization of the joint velocity prevents the solver from computing motions that are too aggressive.
We set~$\bm q_\text{rest}$ to be the first configuration read after starting the controller.

\noindent\textbf{Control regularization cost.}
 The control regularization, achieved by $l_u(\bm x, \bm u) = \left(\bm u - \bm u_\text{rest}(\bm x)\right)^\top Q_u \left(\bm u - \bm u_\text{rest}(\bm x)\right)$, regularizes the controls so that they are not far from $\bm u_\text{rest}(\bm x)$, where $\bm u_\text{rest}(\bm x)$ is a torque that compensates for gravity at the robot configuration~$\bm x$.

\section{Experiments}\label{sec:experiments}
In this section, we first quantitatively evaluate the proposed perception module on the YCBV dataset~\cite{calli2015benchmarking} that contains standardized objects that we also use in the second part of the section for the 6D pose-guided feedback control task on a real Franka Emika Panda robot.
For the implementation of the localizer, we use CosyPose~\cite{labbe2020cosypose} and for the tracker we use ICG~\cite{stoiber2022iterative} unless specified otherwise.

\subsection{Quantitative evaluation of the perception module}\label{sec:exp_benchmark}
\begin{figure}[t]
    \centering
    \includegraphics[width=\linewidth]{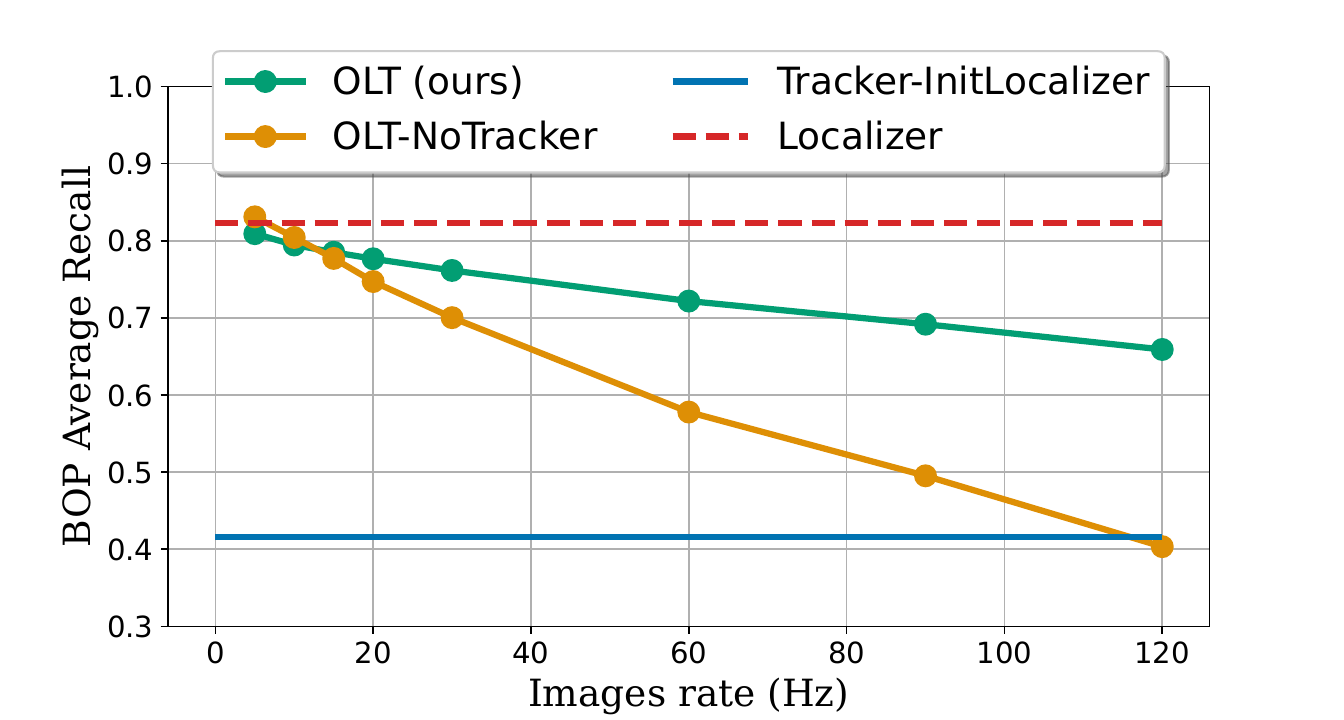}
    \caption{Average recall (higher is better) of BOP metrics \cite{sundermeyer2023bop} measuring the accuracy of 6D pose estimation  of different implementations of object localization and tracking. The comparison was run on the YCBV video dataset replayed at different frequencies on the same hardware.
    }
    \label{fig:ar_f_freq}
\end{figure}

We quantitatively evaluate the new perception module  on the YCBV dataset~\cite{calli2015benchmarking} using the 6D object pose (BOP benchmark) evaluation metrics~\cite{sundermeyer2023bop}.
The YCBV dataset consists of several videos of a moving camera showing a subset of $22$ objects available in the dataset. 
Every frame of the video is annotated with the ground truth poses for all objects visible in the scene.
We use the YCBV dataset because of the availability of the real objects for real-world experiments and of the pre-trained models for the CosyPose~\cite{labbe2020cosypose} object pose estimator. 
We use the BOP toolkit~\cite{sundermeyer2023bop} to compute standard 6D pose error metrics to assess the quality of pose estimates.
The evaluation procedure feeds the images of the input video sequence in order and with a given frequency to the perception module.
The output poses are compared with the ground truth by evaluating \textit{BOP Average Recall} score defined in~\cite{sundermeyer2023bop}.
The results of the evaluation procedure are shown in Fig.~\ref{fig:ar_f_freq} and discussed next.

\begin{figure}[b]
    \centering
    \includegraphics[width=\linewidth]{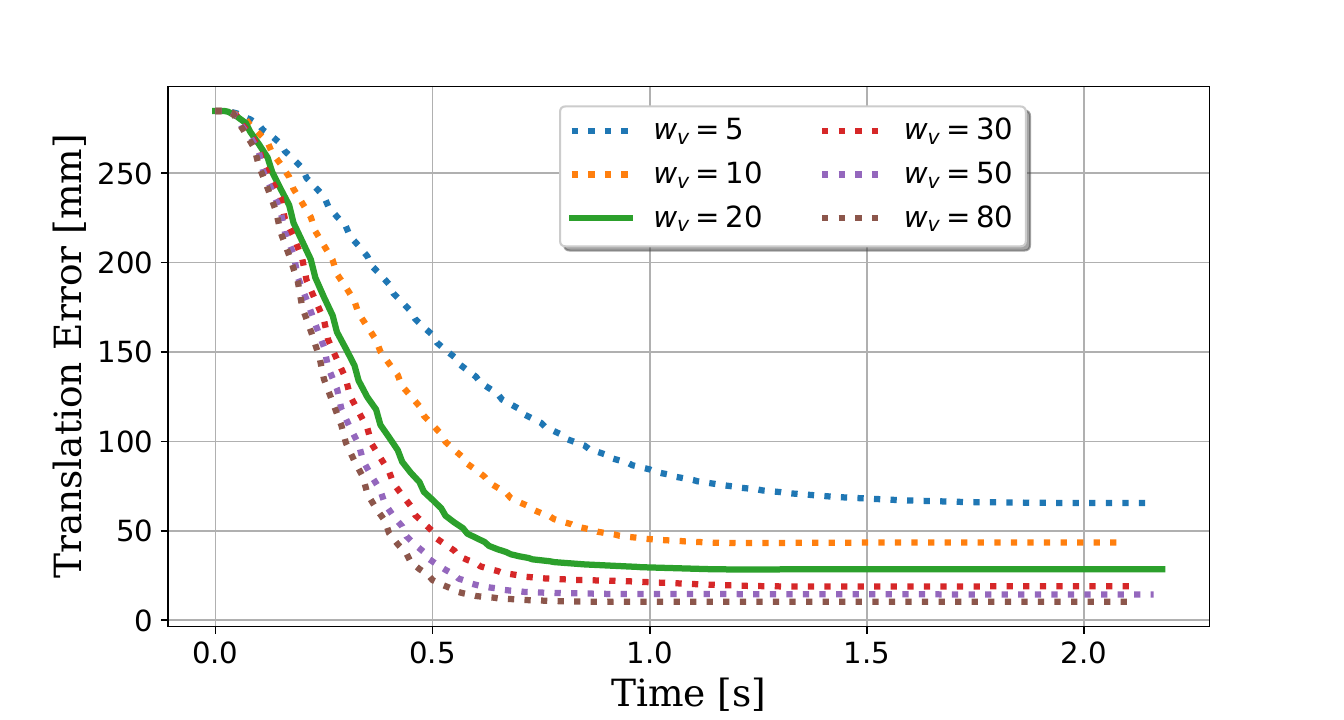}
    \caption{Step-response of the MPC (without the perception) after simulating the target pose rotation by 30~degrees.}
    \label{fig:translation_error}
\end{figure}

\begin{figure*}[t!]
    \centering
    \begin{subfigure}{0.19\linewidth}
        \includegraphics[width=\linewidth]{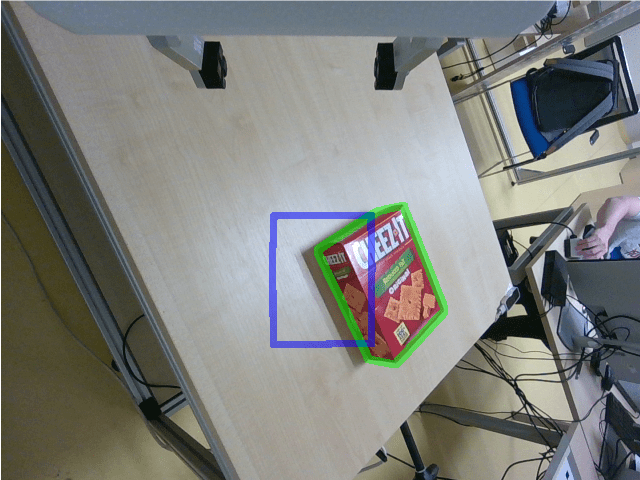}\vspace{1mm}
        % \caption{}
    \end{subfigure}
    \begin{subfigure}{0.19\linewidth}
        \includegraphics[width=\linewidth]{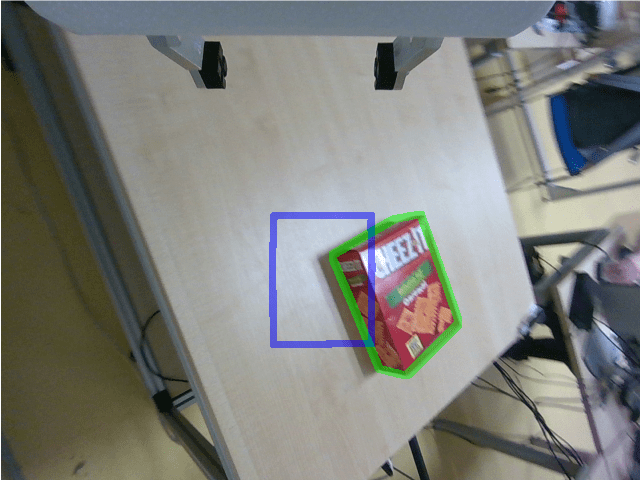}\vspace{1mm}
        % \caption{}
    \end{subfigure}
    \begin{subfigure}{0.19\linewidth}
        \includegraphics[width=\linewidth]{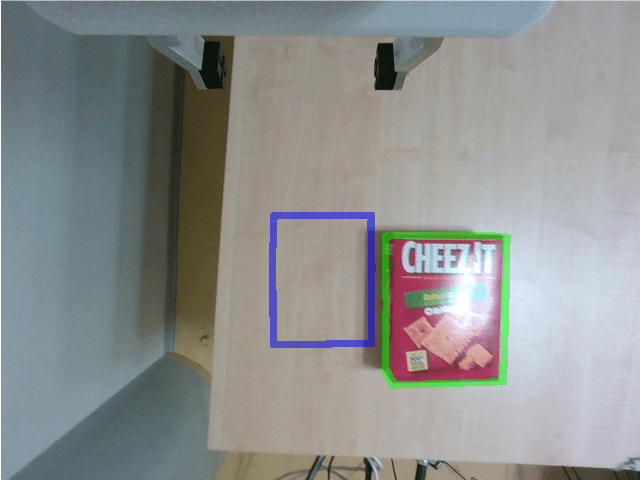}\vspace{1mm}
        % \caption{}
    \end{subfigure}
    \begin{subfigure}{0.19\linewidth}
        \includegraphics[width=\linewidth]{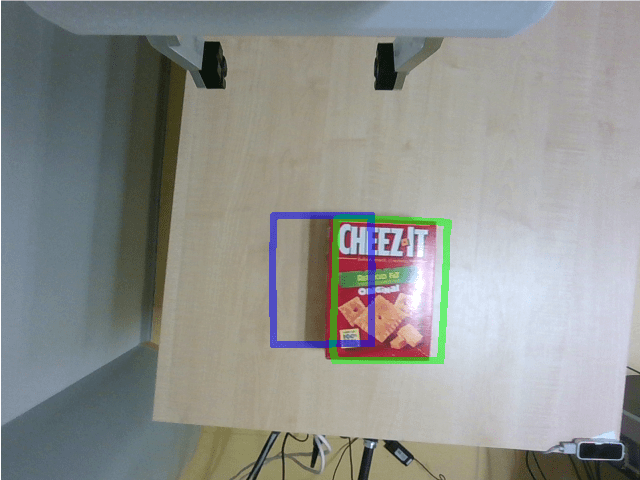}\vspace{1mm}
        % \caption{}
    \end{subfigure}
    \begin{subfigure}{0.19\linewidth}
        \includegraphics[width=\linewidth]{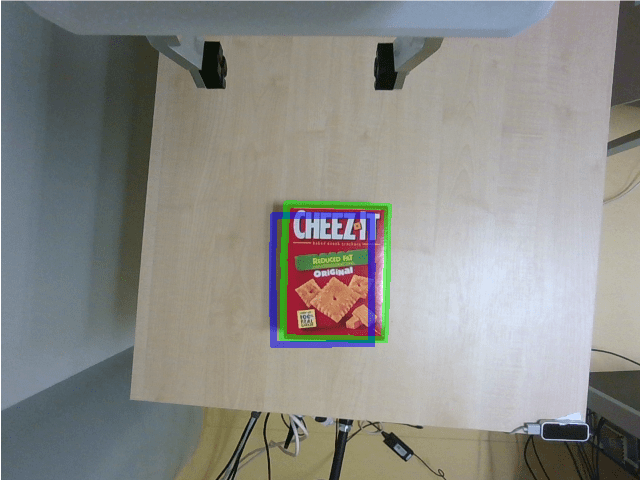}\vspace{1mm}
        % \caption{}
    \end{subfigure}
    %%%%%%%%%%%%%%%%%%%
    \begin{subfigure}{0.19\linewidth}
        \includegraphics[width=\linewidth]{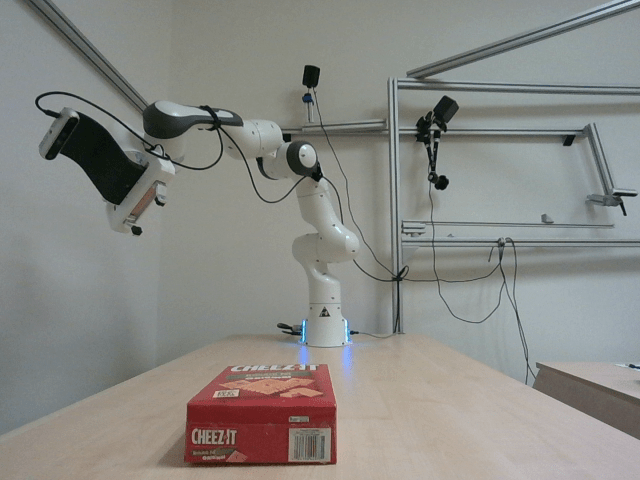}
        % \caption{}
    \end{subfigure}
    \begin{subfigure}{0.19\linewidth}
        \includegraphics[width=\linewidth]{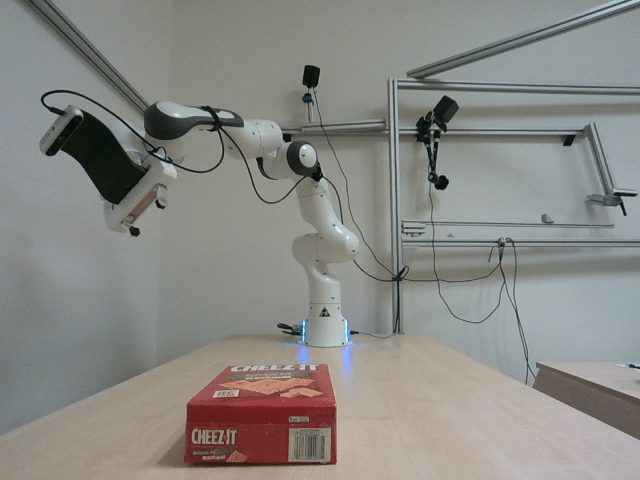}
        % \caption{}
    \end{subfigure}
    \begin{subfigure}{0.19\linewidth}
        \includegraphics[width=\linewidth]{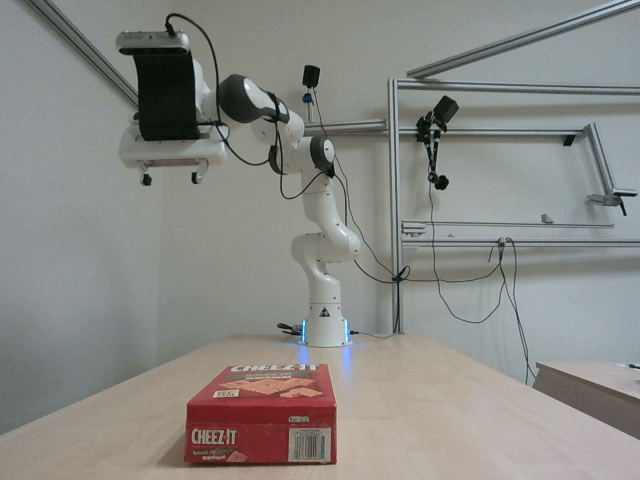}
        % \caption{}
    \end{subfigure}
    \begin{subfigure}{0.19\linewidth}
        \includegraphics[width=\linewidth]{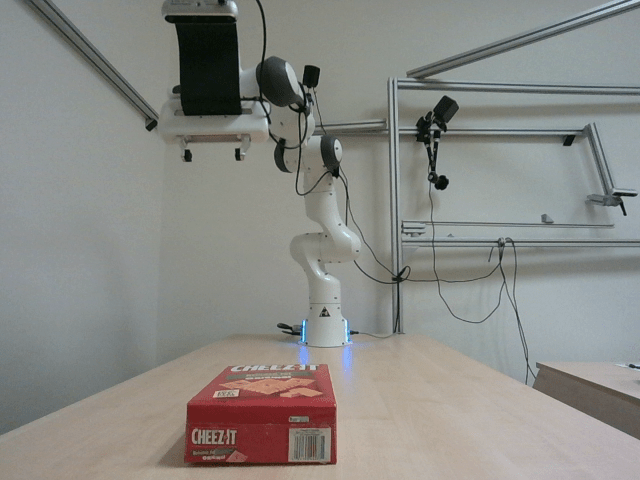}
        % \caption{}
    \end{subfigure}
    \begin{subfigure}{0.19\linewidth}
        \includegraphics[width=\linewidth]{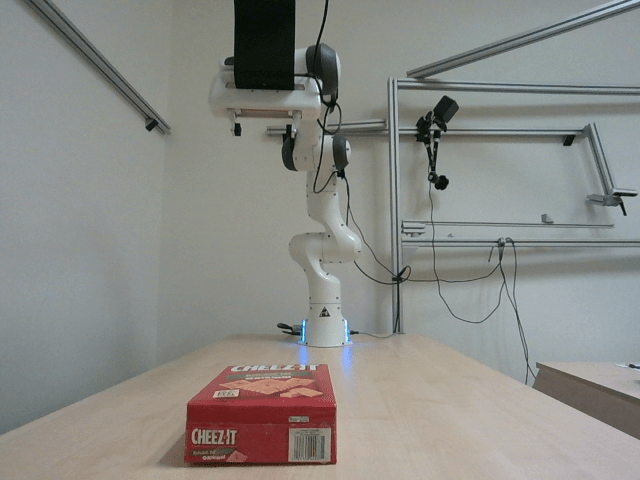}
        % \caption{}
    \end{subfigure}
    \caption{
    \textbf{Visualization of the step-response experiment.}
    The first row shows the images captured by the camera mounted on the robot together with the projection of the pose estimated by our perception module (\textcolor{overviewgreen}{green contour}) and the projection of the target reference pose (\textcolor{overviewblue}{blue contour}).
    The second row shows the images captured by an external camera depicting the motion of the robot.
    The goal of the controller is to move the robot end-effector to a given fixed relative pose w.r.t. the detected object pose.
    The initial configuration of the manipulator is intentionally set away from the target to evaluate the step response of the system.
    The controller brings the camera close to the desired reference pose in 1.5 seconds. {\textbf{Please see the supplementary video for additional examples and experimental analysis.}}
    % Result of a closed loop experiment involving our perception and control modules. The controller is given a fixed 6D reference pose of the object with respect to the camera frame it needs to reach. The initial configuration of the manipulator is intentionally set away from the target 6D pose with respect to the object. The controller brings the camera close to the desired reference pose in 1.5 seconds. \textit{First row}: visualization of the latest estimated 6D pose of the object with respect to the camera (overlaid as a green contour) as well as the target reference 6D object pose (shown as a blue contour). \textit{Second row}: Visualization of the robot movement from an outside point of view.
    }
    \label{fig:vs_expe_cheezit_table}
\end{figure*}

\noindent\textbf{Evaluation baselines.} 
There are two main baselines shown in the plot: 
(i)~\textit{Localizer}, that runs localizer on every frame of the video,
and (ii)~\textit{Tracker-InitLocalizer} that runs the tracker with initialization computed by the localizer on the first frame of the video.
Both baselines are shown as horizontal lines as they were evaluated independently on the frequency of the image stream.
The \textit{Localizer} is introducing high time delay in the system and, therefore, is not suitable for closed-loop control.
However, the results show that the localizer is accurate.
The \textit{Tracker}, on the other hand, runs online with only a small delay, but is not capable of (re-)discovering new or lost object tracks.
Therefore, its average recall is small.

\noindent\textbf{OLT evaluation.}
Our method (\textit{OLT}), lies in between the two baselines.
It runs online with the same delay as \textit{Tracker} but also achieves the \textit{Localizer}'s recall for the low frequencies of the input image stream.
The average recall drops with increasing frequency as the output pose of the localizer is injected into the tracker less frequently.
Asymptotically, for image stream frequency approaching the tracker computation frequency, the OLT's recall would approach performance of pure tracker as the time delay corrector would never \textit{catch-up} in time with the tracker process.

\noindent\textbf{OLT without tracker.}
To assess the influence of the tracker, we perform an ablation study in which we redefine the tracker as identity mapping, \ie $T_k = \ftracker(I_k, T_{k_1}) := T_{k-1}$.
The recall obtained for various image stream frequencies is shown as \textit{OLT-NoTracker} curve in Fig.~\ref{fig:ar_f_freq}.
It can be seen that the influence of the ICG tracker (\ie~OLT (ours)) increases with frequency compared to the identity tracker.
Therefore, the local tracker plays an important role during the computing time of the localizer.
Note that this effect is more pronounced for fast-moving objects which are not present in the YCBV dataset.

\noindent\textbf{Replicability of the results.}
As shown in Fig.~\ref{fig:ar_f_freq}, the average recall of our method depends on the input image stream frequency, as the localizer produces a more accurate 6D pose less often.
Therefore, the computed values also depend on the hardware, and thus are only comparable when run on comparable setups.
We evaluated all methods on the same computer equipped with 12 cores \textit{AMD® Ryzen Threadripper PRO 3945WX} CPU and a \textit{NVIDIA GeForce RTX 3080} GPU.

 \subsection{Visually guided feedback control}
 \label{subsec:feedback}
We experimentally validate the design of our 6D object perception module together with the MPC controller using the following experimental setup.
We use a 7~DoF Franka Emika Panda robot equipped with a RealSense D455 camera attached to its end-effector (eye-in-hand configuration).
The camera mounting w.r.t. end-effector was calibrated.
We configured the camera to produce an RGB video stream at 30 Hz with a 640x480 resolution.
We control the Panda robot in torque-level control mode that requires commands to be sent at 1~kHz frequency.
This justifies the use of the Ricatti Linearization module, which guarantees that a torque command will be computed at this rate.
The OCP is solved with Crocoddyl~\cite{mastalli2020crocoddyl} which uses the efficient robot dynamics implementation from Pinocchio~\cite{carpentier2019pinocchio}. 
To model the dynamics of the robot, we use inertial parameters from~\cite{gaz2019dynamic}.
Computations of the perception module and OCP solving is handled by a computer described in Sec.~\ref{sec:exp_benchmark}.
The 1~kHz Ricatti linearization control loop is computed on another real-time-preempted computer to guarantee the response time requirements of the Panda robot.
The communication between the real-time and the non-real-time computer is implemented using the robotic operating system (ROS)~\cite{Quigley2009ROSAO}.

\noindent\textbf{MPC control evaluation.}
To assess the quality of the MPC control, we perform an experiment on a Panda robot where we analyze the step response of the controller after artificially rotating the target 6D pose by 30~degrees, \ie~the perception module is not used in this experiment.
The evolution of the translation tracking error is shown in Fig.~\ref{fig:translation_error} for different values of tracking weight~$w_v$ (see eq.~\eqref{eq:costs}).
The results confirm that the tracker converges towards the target with steady-state error depending on the tracking weight.
The orientation error (not shown) follows the same pattern.
Based on the results, we have chosen the tracking weight~$w_v$ to be equal to $20$ since the steady-state error is acceptable for our task and a lower weight leads to less aggressive behavior of the controller.
The other weights in the cost were set as $Q_x = \diag(0.3, \ldots, 0.3, 3, \ldots, 3)$ and $Q_u = \diag(0.1, \ldots, 0.1)$.

\noindent\textbf{MPC tracking validation.} We set up a closed-loop robot control experiment shown in Fig.~\ref{fig:vs_expe_cheezit_table} in which the perception and control modules enable us to bring the end-effector of the robot to a desired reference pose with respect to a YCBV object.
Despite the relatively fast motion of the end effector and the presence of specular reflections on the object surface, the tracker is able to maintain an accurate estimation of the object pose throughout the trajectory. 
More examples are presented in the accompanying video.

\section{Conclusion}

Accurate and low-latency object pose estimation is necessary to enable robot interaction with dynamically moving objects, for example, in human-robot handover tasks.
Our work shows that a high-accuracy but slow 6D pose localizer and fast frame-to-frame 6D pose object trackers can be combined to obtain low latency ($<$ 5~ms) pose estimates.
The proposed algorithm has been validated through both (\textit{i})~a quantitative study on a benchmark of common household objects and (\textit{ii})~by developing an MPC-based object pose tracking feedback controller. 
 This work opens up the possibility of visually guided manipulation in 3D dynamic environments, for example, in human-robot collaboration or mobile robot manipulation,
 without the need for fiducial markers or motion capture systems.

% \clearpage
% \balance 
\bibliographystyle{ieeetr}
\bibliography{refs}

\end{document}